\documentclass[runningheads,a4paper]{llncs}

\usepackage{amssymb}
\usepackage{graphicx}

\usepackage{url}
\urldef{\mailsa}\path{frank.glavin@gmail.com,  michael.madden@nuigalway.ie}
\newcommand{\keywords}[1]{\par\addvspace\baselineskip
\noindent\keywordname\enspace\ignorespaces#1}

\begin{document}

\mainmatter  

\title{Analysis of the Effect of Unexpected Outliers in the Classification of Spectroscopy Data}

\titlerunning{Effect of Unexpected Outliers in Spectroscopy Data}

%
%
\author{Frank G. Glavin%
\and Michael G. Madden}
\authorrunning{Frank G. Glavin and Michael G. Madden }

\institute{College of Engineering and Informatics,\\
National University of Ireland, Galway, Ireland.\\
\email{frank.glavin@gmail.com, michael.madden@nuigalway.ie} }

%
%

\maketitle

\begin{abstract}
Multi-class classification algorithms are very widely used, but we argue that they are not always ideal from a theoretical perspective, because they assume all classes are characterized by the data, whereas in many applications, training data for some classes may be entirely absent, rare, or statistically unrepresentative. We evaluate one-sided classifiers as an alternative, since they assume that only one class (the target) is well characterized. We consider a task of identifying whether a substance contains a chlorinated solvent, based on its chemical spectrum. For this application, it is not really feasible to collect a statistically representative set of outliers, since that group may contain \emph{anything} apart from the target chlorinated solvents.  Using a new one-sided classification toolkit, we compare a One-Sided k-NN algorithm with two well-known binary classification algorithms, and conclude that the one-sided classifier is more robust to unexpected outliers.
\end{abstract}

\section{Introduction}

\subsection{One-Sided Classification}
One-sided classification (OSC) algorithms are an alternative to conventional multi-class classification algorithms. They are also referred to as single-class or one-class classification algorithms, and differ in one vital aspect from multi-class algorithms, in that they are only concerned with a single, well-characterized class, known as the target or positive class. Objects of this class are distinguished from all others, referred to as outliers, that consist of \emph{all} the other objects that are not targets. In one-sided classification, training data for the outliers may be either rare, entirely unavailable or statistically unrepresentative.

Over the past decade, several well-known algorithms have been adapted to work with the one-sided paradigm. Tax [1] describes many of these one-sided algorithms and notes that the problem of one-sided classification is generally more difficult than that of conventional classification. The decision boundary in the multi-class case has the benefit of being well described from both sides with appropriate examples from each class being available, whereas the single-class case can only support one side of the decision boundary fully, in the absence of a comprehensive set of counter-examples. While multi-class (including binary or two-class) algorithms are very widely used in many different application domains, we argue that they are not always the best choice from a theoretical perspective, because they assume that all classes are appropriately characterized by the training data. We propose that one-sided classifiers are more appropriate in these cases, since they assume that only the target class is well characterized, and seek to distinguish it from any others. Such problem domains include industrial process control, document author identification and the analysis of chemical spectra.

\subsection{Spectroscopic Analysis}

Raman spectroscopy, which is a form of molecular spectroscopy, is used in physical and analytical chemistry. It involves the study of  experimentally-obtained spectra by using an instrument such as a spectrometer [2]. According to Gardiner [3], Raman spectroscopy is a well-established spectroscopic technique which involves the study of vibrational and rotational frequencies in a system. Spectra are gathered by illuminating a laser beam onto a substance under analysis and are based on the vibrational motion of the molecules which create the equivalent of a chemical fingerprint. This unique pattern can then be used in the identification of a variety of different materials [4].

\subsection{Machine Learning Task}
In this work, we consider the task of identifying materials from their Raman spectra, through the application of both one-sided and multi-class classification algorithms. Our primary focus is to analyse the performance of the classifiers when ``unexpected'' outliers are added to the test sets. The spectra are gathered from materials in pure form and in mixtures. The goal is to identify the presence or absence of a particular material of interest from its spectrum. This task can be seen as an ``open-ended'' problem, as having a statistically representative set of counter-examples for training is not feasible, as has been discussed already.

In particular, we consider the application of separating materials to enable the safe disposal of harmful solvents. Chemical waste that is   potentially hazardous to the environment should be identified and disposed of in the correct manner. Laboratories generally have strict guidelines in place, as well as following legal requirements, for such procedures. Organic solvents can create a major disposal problem in organic laboratories as they are usually water-immiscible and can be highly flammable [5]. Such solvents are generally created in abundance each day in busy laboratories. Differentiating between chlorinated and non-chlorinated organic solvents is of particular importance. Depending on whether a solvent is chlorinated or not will dictate how it is transported from the laboratory and, more importantly, what method is used for its disposal [6]. Identifying and labeling such solvents is a routine laboratory procedure which usually makes the disposal a straightforward process. However, it is not unlikely that the solvents could be accidentally contaminated or inadvertently mislabeled. In such circumstances it would be beneficial to have an analysis method that would correctly identify whether or not a particular solvent was chlorinated.

We have carried out several experiments for this identification using both one-sided and multi-class classification algorithms in order to analyse the effect of adding ``unexpected'' outliers to the test sets.

\section{Related Research}

Madden and Ryder [7] explore the use of standard multi-class classification techniques, in comparison to statistical regression methods, for identifying and quantifying illicit materials using Raman spectroscopy. Their research involves using dimension reduction techniques to select some features of the spectral data and discard all others. This feature selection process is performed by using a Genetic Algorithm. The predictions can then be made based only on a small number of data points. The improvements that can be achieved by using several different predictor models together were also noted. This would come at the cost of increased computation but was shown to provide better results than using just one predictor by itself.

O'Connell {\it et al.} [8] propose the use of Principal Component Analysis (PCA), support vector machines (SVM) and Raman spectroscopy to identify  an analyte\footnote{An analyte is a substance or chemical constituent that is determined in an analytical procedure.} in solid mixtures. In this case, the analyte is acetaminophen, which is a pain reliever used for aches and fevers. They used near-infrared Raman spectroscopy to analyse a total of 217 samples, some of which had the target analyte present, of mixtures with excipients\footnote{An excipient is an inactive substance used as a carrier for the active ingredients of a medication.} of varying weight. The excipients that were included were sugars, inorganic materials and food products. The spectral data was subjected to first derivative and normalization transformations in order to make it more suitable for analysis. After this pre-treatment, the target analyte was then discriminated using Principal Component Analysis (PCA), Principal Component Regression (PCR) and Support Vector Machines. According to the authors, the superior performance of SVM was particularly evident when raw data was used for the input. The importance and benefits of the pre-processing techniques was also emphasized.

Howley [9] uses machine learning techniques for the identification and quantification of materials from their corresponding spectral data. He shows how using Principal Component Analysis (PCA) with machine learning methods, such as SVM, could produce better results than the chemometric technique of Principal Component Regression (PCR). He also presents customized kernels for use with spectral analysis based on prior knowledge of the domain. A genetic programming technique for evolving kernels is also proposed for when no domain knowledge is available.

\section{A Toolkit for One-sided Classification}
In the course of our research, we have developed a one-sided classification toolkit written in Java. It is a command line interface (CLI) driven software package that contains one-sided algorithms that may be chosen by the user at runtime and used to create a new classifier based on a loaded data set and a variety of different adjustable options. Both experiment-specific and classifier parameter options can be set. The toolkit was designed to carry out comprehensive and iterative experiments with minimal input from the user. The resulting classifiers that are generated can be saved and used at a later stage to classify new examples. The user can set up many different runs of an experiment, each differing by an incremented random number seed that shuffles the data for every run before it is broken up into training and testing sets. Results are printed to the screen as they are calculated; these include the classification error, sensitivity, specificity and confusion matrix for each run or individual folds.

\section{Data Sets and Algorithms Used}
\subsection{Primary Data Set}
The primary data set that we used for these experiments was compiled in earlier research, as described by Conroy {\it et al.} [6]. It comprises of 230 spectral samples that contain both chlorinated and non-chlorinated mixtures. According to the authors, the compilation of the data involved keeping the concentrations of the mixtures as close as possible to real life scenarios from industrial laboratories. Twenty five solvents, some chlorinated and some not, were included; these are listed in Table 1.

\begin{table}[h]
\caption{A list of the various chlorinated and non-chlorinated solvents used in the primary data set and their grades.(Source: Conroy {\it et al.} [6])}
\label{Chlorinated and non-chlorinated solvents}
\begin{center}
\begin{tabular}{|c|c||c|c|}
\hline
\multicolumn{1}{|c|}{\bf Solvent}  &\multicolumn{1}{|c||}{\bf Grade} &\multicolumn{1}{|c|}{\bf Solvent} &\multicolumn{1}{|c|}{\bf Grade}
\\ \hline
Acetone         	&HPLC													    &Cyclopentane			  &Analytical\\	
Toluene           &Spectroscopic 										&Acetophenol				&Analytical\\
Cyclohexane				&Analytical \& Spect.  						&n-Pentane					&Analytical\\
Acetonitrile			&Spectroscopic										&Xylene							&Analytical\\
2-Propanol			  &Spectroscopic										&Dimethylformanide	&Analytical\\
1,4-Dioxane				&Analytical \& Spect.							&Nitrobenzene				&Analytical\\
Hexane						&Spectroscopic										&Tetrahydrofuran		&Analytical\\	
1-Butanol					&Analytical \& Spect.							&Diethyl Ether			&Analytical\\	
Methyl Alcohol		&Analytical												&Petroleum Acetate	&Analytical\\
Benzene						&Analytical												&Chloroform					&Analytical \& Spect.\\
Ethyl Acetate			&Analytical												&Dichloromethane		&Analytical \& Spect.\\
Ethanol						&Analytical												&1,1,1-trichloroethane &Analytical \& Spect.\\
\hline
\end{tabular}
\end{center}
\end{table}
\begin{table}[h]
\caption{Summary of chlorinated and non-chlorinated mixtures in the primary data set. (Source: Howley [9])}
\begin{center}
\begin{tabular}{|ccc|c|}
\hline
\multicolumn{1}{|c}{}  &\multicolumn{1}{c}{\bf Chlorinated} &\multicolumn{1}{c}{\bf Non-chlorinated} &\multicolumn{1}{|c|}{\bf Total}\\
\hline
Pure Solvents					&6			&24		&30  \\
Binary Mixtures				&96			&23		&119  \\
Ternary Mixtures			&40			&12		&52    \\
Quaternary Mixtures		&12			&10		&22   \\
Quintary Mixtures			&0			&7		&7   \\
\hline
{\bf Total}						&154		&76		&230  \\
\hline
\end{tabular}
\end{center}
\end{table}

Several variants of the data set were created, which differed only by the labeling of the solvent that was currently assigned as the target class. In each of these variants, all instances not labeled as targets were labeled as outliers. These relabeled data sets were used in the detection of the specific chlorinated solvents: Chloroform, Dichloromethane and Trichloroethane. As an example of the data, the Raman spectrum of pure Chloroform, a chlorinated solvent, is shown in Fig. 1. Other samples from the data set consist of several different solvents in a mixture which makes the classification task more challenging. A final separate data set was created such that all of the chlorinated solvents were labeled as targets. This is for carrying out experiments to simply detect whether a given mixture is chlorinated or not.

\subsection{Secondary Data Set}
For our Scenario 2 experiments (see Section 5.1), we introduce 48 additional spectra that represent outliers that are taken from a different distribution to those that are present in the primary dataset. These samples are the Raman spectra of various laboratory chemicals, and none of them are chlorinated solvents nor are they the other materials that are listed in Table 1. They include materials such as sugars, salts and acids  in solid or liquid state, including Sucrose, Sodium, Sorbitol, Sodium Chloride, Pimelic Acid, Acetic Acid, Phthalic Acid and Quinine.

\begin{figure}[htp]
\centering
\includegraphics[width=0.8\textwidth]{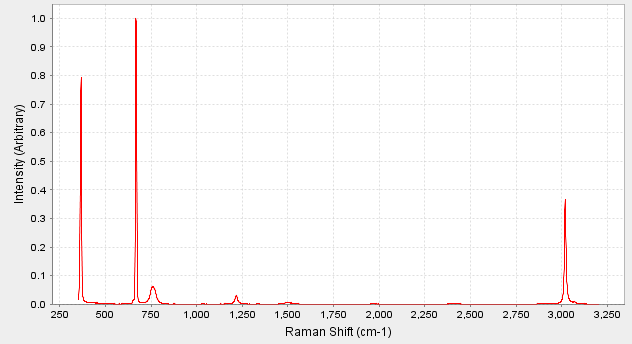}
\caption[Individual spectra of Chloroform]
{\it The Raman Spectrum of a sample of 100\% pure Chloroform} \label{fig:3 spectra}
\end{figure}

\subsection{Algorithms Used}
We carried out the one-sided classification experiments using our toolkit. The conventional classification experiments were carried out using the  Weka [10] machine learning software.

We have chosen a One-Sided k-Nearest Neighbour (OkNN) algorithm and two conventional classification algorithms; namely, k-Nearest Neighbour, that we refer to as Two-Class KNN, and a Support Vector Machine (SVM) that we will refer to as Two-Class SVM.

The OkNN algorithm we use is based on one described by Munroe and Madden [11]. The method involves choosing an appropriate threshold and number of  neighbours to use. The average distance from a test example `A' to its m nearest neighbours is found and this is called `D1'. Then, the average distance of these neighbours to their own respective k nearest neighbours is found and called `D2'. If `D1' divided by `D2' is greater than the threshold value, the test example `A' is rejected as being an outlier. If it is less than the threshold, then it is accepted as being part of the target class.

\section{Description of Experiments}
\subsection{Scenarios Considered}
Two scenarios are described in our experiments, as described next.
\subsubsection{Scenario 1: ``Expected'' Test Data Only.}
In this scenario, the test data is sampled from the same distribution as the training data.  The primary dataset is divided repeatedly into training sets and test sets, with the proportions of targets and outliers held constant at all times, and these 'internal' test sets are used to test the classifiers that are built with the training datasets.

\subsubsection{Scenario 2: ``Unexpected'' and ``Expected'' Test Data.}
In this scenario, we augment each test dataset with the 48 examples from the secondary data set that are \emph{not} drawn from the same distribution as the training dataset. Therefore, a classifier trained to recognise any chlorinated solvent should reject them as outliers. However, these samples represent a significant challenge to the classifiers, since they violate the standard assumption that the test data will be drawn from the same distribution as the training data; it is for this reason that we term them ``unexpected''.

This second scenario is designed to assess the robustness of the classifiers in a situation that has been discussed earlier, whereby in practical deployments of classifiers in many situations, the classifiers are likely to be exposed to outliers that are not drawn from the same distribution as training outliers. In fact, we contend that over the long term, this is inevitable: if we know \emph{a priori} that the outlier class distribution is not well characterized in the training data, then we must accept that sooner or later, the classifier will be exposed to data that falls outside the distribution of the outlier training data.

It should be noted that this is different from \emph{concept drift}, where a target concept may change over time; here, we have a static concept, but over time the weaknesses of the training data are exposed. Of course, re-training might be possible, if problem cases can be identified and labeled correctly, but we concern ourselves with classifiers that have to maintain robust performance without re-training.

\subsection{Experimental Procedure}
The data sets, as described earlier, were used to test the ability of each algorithm in detecting the individual chlorinated compounds. This involved three separate experiments for each algorithm, to detect Chloroform, Dichloromethane and Trichloroethane. A fourth experiment involved detecting the presence of any chlorinated compound in the mixture.

All spectra were first normalized. A common method for normalizing a dataset is to recalculate the values of the attributes to fall within the range of zero to one. This is usually carried out on an attribute-by-attribute basis and ensures that certain attribute values, which differ radically in size from the rest, don't dominate in the prediction calculations. The normalization carried out on the spectral data is different to this in that it is carried out on an instance-by-instance basis. Since each attribute in an instance is a point on the spectrum, this process is essentially rescaling the height of the spectrum into the range of zero to one.

For each experiment, 10 runs were carried out with the data being randomly split each time into 67\% for training and 33\% for testing. The splitting procedure from our toolkit ensured that there was the same proportion of targets and outliers in the training sets as there was in the test sets. The same data set splits were used for the one-sided classifier algorithms and the Weka-based algorithms, to facilitate direct comparisons.

A 3-fold internal cross validation step was used with all the training sets, to carry out parameter tuning. A list of parameter values was passed to each classification algorithm and each, in turn, was used on the training sets, in order to find the best combination that produced the smallest error estimate. It must be emphasized that we only supplied a small amount of different parameters for each algorithm and that these parameters used were the same for all of the four variants of the data set. The reason for this was that our goal was not to tune and identify the classifier with the best results overall but to notice the change in performance when ``unexpected'' outliers were added to the test set.

For the One-Sided kNN algorithm, the amount of nearest neighbours (m) and the amount of their nearest neighbours (k) was varied between 1 and 2. The threshold values tried were 1, 1.5 and 2. The distance metric used was Cosine Similarity. For the Weka experiments, the Two-Class kNN approach tried the values 1,2 and 3 for the amount of nearest neighbours. The Two-Class SVM varied the complexity parameter C with the values of 1,3 and 5. The default values were used for all of the other Weka parameters.

\subsection{Performance Metric}
The error rate of a classification algorithm is the percentage of examples from the test set that are incorrectly classified. We measure the average error rate of each algorithm over the 10 runs to give an overall error estimate of its performance.
With such a performance measure being used, it is important to know what percentage of target examples were present in each variant of the data set. This information is listed in Table 3 below.

\begin{table}
\caption{Percentage of target examples in each variant of the \emph{primary} data set}
\begin{center}
\begin{tabular}{|c||c|c|c|}
\hline
\multicolumn{1}{|c||}{Dataset}  &\multicolumn{1}{|c|}{\bf Targets} &\multicolumn{1}{|c|}{\bf``Expected'' Outliers} &\multicolumn{1}{|c|}{\bf Target Percent}\\
\hline{\bf Chlorinated or not}					&154			&76  	&66.95\% \\
\hline{\bf Chloroform}									&79				&151 	&34.34\% \\
\hline{\bf Dichloromethane}							&60				&170 	&26.08\% \\
\hline{\bf Trichloroethane}							&79				&151 	&34.34\% \\
\hline
\end{tabular}
\end{center}
\end{table}

\section{Results and Analysis}
The results of the experiments carried out are listed in Table 4, Table 5, and Table 6 below. Each table shows the overall classification error rate and standard deviation (computed over 10 runs) for each algorithm, for both of the scenarios that were tested.

It can be seen that while the conventional multi-class classifiers perform quite well in the first scenario, their performance quickly begins to deteriorate once the ``unexpected'' outliers are introduced in Scenario 2. The One-Sided kNN's performance is generally worse than the multi-class approach in Scenario 1. As described in Section 1.1, the decision boundary for the multi-class classifiers have the benefit of being well supported from both sides with representative training examples from each class. In such a scenario, the multi-class algorithms essentially have more information to aid the classification mechanism and, therefore, would be expected to out-perform the one-sided approach.

In detecting whether or not a sample is chlorinated, the average error rate of the Two-Class kNN increased by 28.87\% and the Two-Class SVM increased by 33.57\% in Scenario 2. In contrast with the two-class classifiers, the One-Sided kNN is seen to retain a consistent performance and the error is only increased by 0.14\%. When the algorithms are detecting the individual chlorinated solvents, the same pattern in performance can be seen. The multi-class algorithms' error rates increase, in some cases quite radically, in the second scenario. The One-Sided kNN manages to remain at a more consistent error rate and, in the case of Chloroform and Dichloromethane, the overall error rate is reduced somewhat.

It should be noted that our experiments are not concerned with comparing the relative performances of a one-sided classifier versus the multi-class classifiers. Rather, we analyse the variance between the two scenarios for each individual classifier and demonstrate the short-comings of the multi-class approach when it is presented with ``unexpected'' outliers. Our results demonstrate the one-sided classifier's ability to robustly reject these outliers in the same circumstances.
\begin{table}
\caption{Overall average error rate for two-class kNN in both scenarios}
\begin{center}
\begin{tabular}{|c||c|c|}
\hline
\multicolumn{1}{|c||}{Two-Class kNN}  &\multicolumn{1}{|c|}{{\bf Scenario 1.}}&\multicolumn{1}{|c|}{\bf Scenario 2.}\\
\hline &Error \% (std. dev.)&Error \% (std. dev.) \\
\hline{\bf Chlorinated or not}					&6.49 (2.03)	&35.36 (3.65) \\
\hline{\bf Chloroform}									&22.59 (6.93)	&39.44 (7.37) \\
\hline{\bf Dichloromethane}							&11.94 (4.89)	&16.24 (3.49)\\
\hline{\bf Trichloroethane}							&23.24 (5.10)	&25.68 (4.27) \\
\hline
\end{tabular}
\end{center}
\end{table}
\begin{table}
\caption{Overall average error rate for two-class SVM in both scenarios}
\begin{center}
\begin{tabular}{|c||c|c|}
\hline
\multicolumn{1}{|c||}{Two-Class SVM}  &\multicolumn{1}{|c|}{{\bf Scenario 1.}} &\multicolumn{1}{|c|}{\bf Scenario 2.}\\
\hline &Error \% (std. dev.)&Error \% (std. dev.) \\
\hline{\bf Chlorinated or not}					&4.67 (1.95)		&38.24 (2.19)\\
\hline{\bf Chloroform}									&11.68 (4.01)		&37.2 (2.39)\\
\hline{\bf Dichloromethane}							&8.70 (4.37)		&11.68 (3.52)\\
\hline{\bf Trichloroethane}							&11.03 (3.47)		&30.08 (2.50)\\
\hline
\end{tabular}
\end{center}
\end{table}
\begin{table}
\caption{Overall average error rate for one-sided kNN in both scenarios}
\begin{center}
\begin{tabular}{|c||c|c|}
\hline
\multicolumn{1}{|c||}{One-Sided kNN}  &\multicolumn{1}{|c|}{{\bf Scenario 1.}} &\multicolumn{1}{|c|}{\bf Scenario 2.}\\
\hline &Error \% (std. dev.)&Error \% (std. dev.) \\
\hline{\bf Chlorinated or not}					&10.90 (4.5)			&11.04 (4.7) \\
\hline{\bf Chloroform}									&26.10 (3.43)			&18.32 (3.16) \\
\hline{\bf Dichloromethane}							&12.98 (3.23)			&9.36 (2.84) \\
\hline{\bf Trichloroethane}							&20.77 (3.46)			&21.04 (5.07) \\
\hline
\end{tabular}
\end{center}
\end{table}

\section{Conclusions and Future Work}
Our research demonstrates the potential drawbacks of using conventional multi-class classification algorithms when the test data is taken from a different distribution to that of the training samples. We believe that for a large number of real-world practical problems, one-sided classifiers should be more robust than multi-class classifiers, as it is not feasible to sufficiently characterize the outlier concept in the training set. We have introduced the term ``unexpected outliers'' to signify outliers that violate the standard underlying assumption made by multi-class classifiers, which is that the test set instances are sampled from the same distribution as the training set instances. We have shown that, in such circumstances, a one-sided classifier can prove to be a more capable and robust alternative.
Our future work will introduce new datasets from different domains and also analyse other one-sided and multi-class algorithms.

\subsubsection*{Acknowledgments.}
The authors are grateful for the support of Enterprise Ireland under Project CFTD/05/222a. The authors would also like to thank Dr. Abdenour Bounsiar for his help and valuable discussions, and Analyze IQ Limited for supplying some of the Raman spectral data.

\end{document}